\pdfoutput=1
\documentclass[runningheads]{llncs}
\usepackage{graphicx}
\usepackage{amsmath,amssymb} 
\usepackage[hyperindex,breaklinks]{hyperref}
\usepackage{color}
\usepackage{listings}
\usepackage{xcolor}

\definecolor{codegreen}{rgb}{0,0.6,0}
\definecolor{codegray}{rgb}{0.5,0.5,0.5}
\definecolor{codepurple}{rgb}{0.58,0,0.82}
\definecolor{backcolour}{rgb}{0.95,0.95,0.92}

\lstdefinestyle{mystyle}{
    backgroundcolor=\color{backcolour},   
    commentstyle=\color{codegreen},
    keywordstyle=\color{magenta},
    numberstyle=\tiny\color{codegray},
    stringstyle=\color{codepurple},
    basicstyle=\ttfamily\footnotesize,
    breakatwhitespace=false,         
    breaklines=true,                 
    captionpos=b,                    
    keepspaces=true,                 
    numbers=left,                    
    numbersep=5pt,                  
    showspaces=false,                
    showstringspaces=false,
    showtabs=false,                  
    tabsize=2
}

\usepackage[width=122mm,left=12mm,paperwidth=146mm,height=193mm,top=12mm,paperheight=217mm]{geometry}
\begin{document}

\pagestyle{headings}
\mainmatter
\def\ECCV16SubNumber{***}  

\title{Super-Selfish: 
Self-Supervised Learning on Images with PyTorch} 

\titlerunning{Super-Selfish}

\author{Nicolas Wagner, Anirban Mukhopadhyay} 
\institute{Technical University of Darmstadt}

\maketitle

\begin{abstract}
Super-Selfish is an easy to use PyTorch framework for image-based self-supervised learning. Features can be learned with 13 algorithms that span from simple classification to more complex state of the art contrastive pretext tasks. The framework is easy to use and allows for pre-training any PyTorch neural network with only two lines of code. Simultaneously, full flexibility is maintained through modular design choices. The code can be found at \url{https://github.com/MECLabTUDA/Super_Selfish} and installed using \texttt{pip install super-selfish}.
\end{abstract}

\section{Algorithms}\label{sec:algos}
Super-Selfish features 13 algorithms categorized into predictive (patch-based), predictive (autoencoding), generative, and contrastive ones. We give a brief overview of contrastive methods as those represent the state of the art in 2020.

\subsection{Contrastive}
Contrastive algorithms all rely on the same idea and differ only in architectural changes. Commonly, a low dimensional embedding is learned for each instance. The trick is to enforce the embedding to be solely based upon the content and not on other potentially meaningless cues (e.g. chromatic aberration). In one way or another, the trained network sees different views of the same image as positive examples and views of other images as negative examples. The network is then trained to output similar embeddings for positive examples and greatly different embeddings for negative examples\footnote{Although BYOL \cite{grill2020bootstrap} states that no negative examples are used this does not seem to be right \url{https://untitled-ai.github.io/understanding-self-supervised-contrastive-learning.html}.} measured by a contrastive loss (e.g. softmax of angular embedding distances).  To the best of our knowledge, there is no fair and unified comparison between methods as those rely on enormous computational power and do not focus on resource constraint university labs. Even worse, to a great extent, contrastive algorithms rely on choosing the right image augmentations and sophisticated training engineering. Contrastive algorithms can be distinguished as follows:

\subsubsection{Unsupervised Feature Learning via Non-Parametric Instance Discrimination}
 \cite{WuXYL18} (ID) is close to the general contrastive idea described above. The first embedding of an image is retrieved by passing it through a neural network. A positive example is retrieved from a memory of older neural network evaluations of the same image. Likewise, negative examples are also extracted from the same memory. Originally, no image augmentations are applied. However, by default, Super-Selfish conducts the same transformations as Contrastive Predictive Coding \cite{CPC}.
 
\subsubsection{Data-Efficient Image Recognition with Contrastive Predictive Coding} 
\cite{CPC} (CPC) splits one image into many patches. The neural network is not only trained to embed each patch but also to predict the embedding of a patch from previous (i.e. above, below, etc.) ones. The positive example for a prediction is the actual embedding of a future patch whereas negative examples are randomly drawn embeddings of patches from other images in the same batch. Here, different projections into the embedding space (i.e. different MLPs on top of the feature extracting backbone) may be used for the prediction and the positive and negative examples, respectively. CPC uses a suite of strong image augmentations.

\subsubsection{Improved Baselines with Momentum Contrastive Learning}
\cite{mom} (MoC) is, in turn, similar to ID. The significant differences are the use of a momentum encoder as well as replacing the memory bank with a queue (FIFO). An image is first embedded using the neural network under optimization. Afterward, a positive example is constructed using the momentum encoder. The momentum encoder is built by a moving average of the neural network parameters. The positive examples of a batch are then appended to the queue and used as negative examples for the following batches. MoC uses its own suite of data augmentations. Further, a temperature parameter is applied to control the distribution concentration of the contrastive loss.

\subsubsection{Contrastive Multiview Coding}
\cite{cmc} (CMC) extends ID by transforming images into the Lab color space. Subsequently, embeddings are either created from the L channel or the ab channels. Positive and negative examples are then retrieved from embedding the missing channels. Both L and ab have their own backbone networks as well as two different projection heads like in CPC. After self-supervision Super-Selfish only provides the features of the L channel as those of the ab channels should be close anyway. Further, we apply the image augmentations of CPC.

\subsubsection{Bootstrap Your Own Latent}
\cite{grill2020bootstrap} (BYOL) extends the idea of the momentum encoder.  However, in contrast to MoC, no negative examples are used. This is achieved by adding an additional projection head to the actual network that is not a constituent of the momentum encoder. Again, a specific set of image augmentations is used.

\subsubsection{Self-Supervised Learning of Pretext-Invariant Representations}
\cite{pirl} uses a memory bank like ID, applies two different projection heads, and adds a temperature parameter like MoC, but adds a significantly different data augmentation. The defining challenge for PIRL is to solve a Jigsaw Puzzle. Here, an image is cut into 3x3 patches which are randomly shuffled (more precisely, a random permutation of the 1000 permutations with the highest hamming distances is drawn). Each patch is processed with the strong CPC augmentation suite. Further, to avoid border cues, each patch is randomly cropped and resized to the original size. Contrary to the original implementation, we pass the shuffled image at once and not each patch separately. Due to the cropping and way stronger image augmentations than actually applied, this should be a negligible design choice in favor of simplicity and efficiency.

\subsection{Predictive (Patch-Based)}
We further implement three other algorithms that work on image patches. First, ExemplarNet \cite{exem}, which considers various views of the same image patch as a class. The task of ExemplarNet is to predict the corresponding class. Second, RotateNet \cite{rotate}, is quite similar to ExemplarNet but only considers 4 different rotations as different classes. Finally, Jigsaw Puzzle \cite{puzzle}, predicts the permutation of a jigsaw puzzle. PIRL is mainly motivated by \cite{puzzle}.
\subsection{Predictive (Autoencoding)}
The predictive algorithms are supplemented by autoencoding algorithms. The most prominent is the Denoising-Autoencoder \cite{denoising} which aims to recover pixels that are corrupted by noise. An evolution of the Denoising-Autoencoder is the Context-Autoencoder \cite{context} that erases multiple crops from an image and tries to recover those. The Context-Autoencoder applies an additional adversarial loss to solve this task. We also implement the Splitbrain-Autoencoder \cite{split} which splits an image into L and ab channels. The task is to predict the ab channels from the L channel and vice versa. The Splitbrain-Autoencoder is related to CMC.

\subsection{Generative}
The BiGAN \cite{bi} is a generative approach to self-supervised learning. It aims to learn the inverse mapping of the generator for extracting features of an image.
\section{Usage}
\subsection{Training}
Training is as easy as:
\begin{lstlisting}[language=Python][language=Python]
# Choose supervisor
supervisor = RotateNetSupervisor(train_dataset) # .to('cuda')

# Start training
supervisor.supervise(lr=1e-3, epochs=50, batch_size=64, name="store/base", pretrained=False)

\end{lstlisting}
\subsection{Best Practice}
General-purpose self-supervision algorithms are mostly evaluated on ImageNet. However, datasets and tasks may differ in terms of resolution and complexity. From our experience, the key to successfully apply self-supervision is to tune the complexity of the pretext task. Further, the state of the art contrastive approaches greatly rely on huge batch sizes that require Multi-GPU hardware setups. If not applicable, MoC is typically still performing well due to the queued structure. Summarizing, the evaluation focus of self-supervised learning on ImageNet may result in a misleading expectation of transferability. All approaches need to be adapted to the dataset and task at hand.

\subsection{Feature Extraction and Transfer}
The model is automatically stored if the training ends after the given number of epochs or the user manually interrupts the training process.  
If not directly reused in the same run, any model can be loaded with:

\begin{lstlisting}[language=Python]
supervisor = RotateNetSupervisor().load(name="store/base")
\end{lstlisting}
The feature extractor is retrieved using:
\begin{lstlisting}[language=Python]
# Returns the backbone network i.e. nn.Module
backbone_network = supervisor.get_backbone()
\end{lstlisting}
If you want to easily add new prediction heads you can create a CombinedNet:
\begin{lstlisting}[language=Python]
CombinedNet(backbone_network, nn.Module(...)) 
\end{lstlisting}

\subsection{Flexibility}
Although training is as easy as writing two lines of code, Super-Selfish provides maximum flexibility. Any supervisor can be directly initialized with the corresponding hyperparameters. By default, the hyperparameters from the respective papers are used. Similarly, the high-level backbone as well as prediction head architectures are by default those of the papers but can be customized as follows:
\begin{lstlisting}[language=Python]
supervisor = RotateNetSupervisor(train_dataset, backbone=nn.Module(...), predictor=nn.Module(...)) # .to('cuda')
\end{lstlisting}
For individual parameters see the full documentation.  

The training can be governed by the learning rate, the used optimizer, the batch size, whether to shuffle training data, and a learning rate schedule. Polyak averaging is soon to be added.
\begin{lstlisting}[language=Python]
def supervise(self, lr=1e-3, optimizer=torch.optim.Adam, epochs=10, batch_size=32, shuffle=True,
                  num_workers=0, name="store/base", pretrained=False, lr_scheduler=lambda optimizer: torch.optim.lr_scheduler.StepLR(optimizer, step_size=100, gamma=1.0))
\end{lstlisting}
The supervise method of any Superviser is split into 5 parts such that functionalities can be easily updated/changed through overloading.
\begin{lstlisting}[language=Python]
# Loading of pretrained weights and models
def _load_pretrained(self, name, pretrained)
# Initialization of training specific objects
def _init_data_optimizer(self, optimizer, batch_size, shuffle, num_workers, collate_fn, lr, lr_scheduler)
# Wraps looping over epochs, batches. Takes care of visualizations and logging.
def _epochs(self, epochs, train_loader, optimizer, lr_scheduler)
# Implements one run of a model and other forward calculations
def _forward(self, data)
# Takes care of updating the modle, lr scheduler, ...
def _update(self, loss, optimizer, lr_scheduler)
\end{lstlisting}

\subsection{Remarks}
If not precisely stated in a paper, we use the CPC image augmentations. Some augmentations or implementation details may be different from the original papers as we aim for a comparable unified framework. Generally, we use an EfficientNet\footnote{\url{https://github.com/lukemelas/EfficientNet-PyTorch}} implementation as the default backbone/feature extractor. Precisely, we use a customized version that can be switched from batch norm to layer norm. Super Selfish is constructed for a $225 \times 225$ resolution but can be adapted with little effort. A responsible architecture is to follow soon. Please feel free to open an issue regarding bugs and/or other algorithms that should be added or contact the authors.
\clearpage

\bibliographystyle{splncs}
\bibliography{egbib}
\end{document}